\documentclass[runningheads]{llncs}
\usepackage{graphicx}

\usepackage{url}
\usepackage{cite}
\usepackage{amsmath,amssymb,amsfonts}
\usepackage{algorithmicx}
\usepackage{algorithm}
\usepackage{algpseudocode}
\usepackage{booktabs}
\usepackage{float}
\usepackage{multicol}
\usepackage{enumitem}
\usepackage{textcomp}
\usepackage{multirow}
\usepackage{xcolor}
\usepackage{adjustbox}
\usepackage[english]{babel}
\usepackage[utf8]{inputenc}
\usepackage[T1]{fontenc}
\usepackage{adjustbox}
\newcommand{\comment}[1]{}
\def\BibTeX{{\rm B\kern-.05em{\sc i\kern-.025em b}\kern-.08em
    T\kern-.1667em\lower.7ex\hbox{E}\kern-.125emX}}

\begin{document}
\title{Portuguese FAQ for Financial Services}
\author{Paulo Finardi\inst{1} \and
Wanderley M. Melo\inst{2} \and
Edgard D. Medeiros Neto\inst{2} \and
Alex F. Mansano\inst{1} \and
Pablo B. Costa\inst{1} \and
Vinicius F. Caridá\inst{1}}
\authorrunning{Finardi et al.}
\institute{MaLS Data Science Team - Digital Customer Service \\
Itaú Unibanco, São Paulo, Brazil\\
\email{{\{paulo.finardi, pablo.costa, alex.mansano, vinicius.carida\}@itau-unibanco.com.br}}\\
 \and
Department of Strategic Management and Specialized Supervision \\ Central Bank of Brazil, Fortaleza, Brazil \\
\email{\{wanderley.melo, edgard.medeiros\}@bcb.gov.br}}
\maketitle              % typeset the header of the contribution
\begin{abstract}

%reescrevi e re-arrangei algumas frases.

Scarcity of domain-specific data in the Portuguese financial domain has disfavored the development of Natural Language Processing (NLP) applications. To address this limitation, the present study advocates for the utilization of synthetic data generated through data augmentation techniques. The investigation focuses on the augmentation of a dataset sourced from the Central Bank of Brazil FAQ, employing techniques that vary in semantic similarity. Supervised and unsupervised tasks are conducted to evaluate the impact of augmented data on both low and high semantic similarity scenarios. Additionally, the resultant dataset will be publicly disseminated on the Hugging Face Datasets platform, thereby enhancing accessibility and fostering broader engagement within the NLP research community.

%The scarcity of financial domain data in Portuguese is a issue for NLP applications. Synthetic data generation with data augmentation techniques is an alternative to circumvent the short data problem. In this work, we explore a Central Bank of Brazil FAQ dataset with Data Augmentation techniques. Furthermore, we performed supervised and unsupervised tasks to measure the effects of augmented data with low and high semantic similarity. We will also make the public dataset available on Hugging Face Datasets, in order to make it more accessible.

\end{abstract}
\section{Introduction}

The application of deep learning methodologies in contexts characterized by limited resources has gained prominence across various domains within Natural Language Processing (NLP) and Natural Language Generation (NLG) \cite{surveyNLP, surveyNLG}. Addressing the challenges posed by low-resource scenarios, transfer learning techniques have been widely employed \cite{trasfer-learning_1, trasfer-learning_2} in Information Retriever (IR) tasks \cite{faq_intro1, faq_intro2}. Despite the efficacy of such approaches, the inherent scarcity of data remains a significant impediment, particularly when confronted with critical domains such as finance. This scarcity is further exacerbated by the private and restricted nature of financial data, which poses challenges in terms of accessibility.

%Ideally, to be able to develop Natural Language Processing (NLP) solutions some minimal amount of data is needed. Transfer learning techniques \cite{trasfer-learning_1, trasfer-learning_2} help to solve this issue. However, data scarcity directly affects model performance. The lack of data is accentuated when the data domain is critical, such as finance. In general, financial data is private and difficult to access. In particular, the Portuguese language has few resources NLP, as pointed out in several works in the literature \cite{petrov-etal-2012-universal, ramesh-sankaranarayanan-2018-neural}

In response to the challenge posed by the scarcity of data in the Portuguese financial domain, this study aims to contribute constructively by utilizing publicly accessible Frequently Asked Question (FAQ) data obtained from the Central Bank of Brazil (BACEN). The obtained dataset will be made available through the Hugging Face Datasets platform \cite{datasets}. The motivation behind employing deep learning methodologies and Information Retrieval (IR) techniques on FAQ data is rooted in the pursuit of identifying an optimal synergy between the model and the data, thereby culminating in an enhanced solution for end-users. The present paper is dedicated to an exhaustive examination of the FAQ dataset, incorporating the application of Data Augmentation (DA) techniques for both unsupervised and supervised Natural Language Processing (NLP) tasks. The main objectives of this research paper are:

%In order to positively contribute to the solution of this problem, we explore Frequently Asked Question (FAQ) public data from the Central Bank of Brazil (BACEN) and will make it available Hugging Face (HF) Datasets \cite{datasets}.  Exploring FAQ data with AI aims to find the best combination of model and data that provides the optimal solution for an end user. Providing a correct answer to the user from an IR system is still a challenge \cite{faq_intro1, faq_intro2}. This paper study the FAQ dataset with Data Augmentation (DA) techniques for unsupervised and supervised NLP tasks, our goals with this paper are: 

\begin{itemize}
    \item Study of public data from the FAQ of BACEN and availability of the data in the HF Datasets;
    \item exploration of DA techniques;
    \item Evaluation of IR tasks and text classification with application to financial services.
\end{itemize}

% Acho que na introdução caberia um pouco do texto do artigo do JP Morgan (https://www.jpmorgan.com/technology/technology-blog/synthetic-data-for-real-insights)

\section{BACEN FAQ dataset}

The public dataset from the Frequently Asked Questions of the Central Bank of Brazil (BACEN FAQ) is accessible through the following website: \url{https://dadosabertos.bcb.gov.br}. The BACEN FAQ dataset comprises roughly 2,000 samples of question-answer pairs represented as ${(q, a)}$. Each entry in this dataset includes not only the question and corresponding answer, but also a broader question category, herein referred to as the macro category. This macro category serves to contextualize the origin of the question, providing insights into the thematic domain of inquiry. For instance, questions of a more generic nature, such as “What is this?” and “How do I access it?” are examples where the macro category aids in refining the subject matter. 

%The BACEN FAQ data is public and it can be obtained through the website: \url{https://dadosabertos.bcb.gov.br}. A FAQ consists of pairs of questions and answers $\{(q, a)\}$, in addition to the question and answer pair, this data also carries a macro question category/subject column. This feature is necessary to understand "where" the question was asked. For example, there are generic questions like "{\itshape What is it?}", "{\itshape How do I access?}" where  without the macro subject becomes too much generic. We made a simple data clean, removing unanswered questions and we get as a final shape 1855 examples.
%The BACEN FAQ data is a public data that and it can be obtained through the website: 

%\subsection{Main FAQ Features} ao meu ver subsções só devem ser usdas se existirem mais de uma.

%Each of the question and answer pairs in the final dataset is categorized into 242 different categories. For example, for the Out-Of-Domain (OOD) category, 289 examples are grouped. Disregarding the OOD category, the three categories that group the largest number of examples are: \textit{Credit Card and Revolving Credit}, \textit{Registrato} and\textit{Real Estate Credit}, all with less than 30 examples each.

The final version of the dataset comprises question-answer pairs categorized into 242 distinct categories. For instance, within the Out-Of-Domain (OOD) classification, 289 instances are encompassed. Excluding the OOD category, the three categories with the highest aggregation of instances are: \textit{Cartão de Crédito e Crédito Rotativo}, \textit{Registrato}, and \textit{Crédito Imobiliário}, each containing fewer than 30 examples.

The Table~\ref{tab:feats} summarizes the main characteristics of the dataset. The average question length is 12 words, with 45 questions containing fewer than 5 words. In contrast, responses exhibit an average length of 78 words. We split the data as usual with 70/30 holdout.

%The dataset has three columns: question, category and answer. There are 242 categories, including the Out-Of-Domain (OOD) category which has 289 examples. Disregarding the OOD, the three most popular categories are: \textit{Cartão de Crédito e Crédito Rotativo}, \textit{Registrato} and \textit{Crédito Imobiliário}, all with less than 30 examples each. The questions have an average length of 12 words, and there are 45 questions with less than 5 words. The average length of the number of words in the answers is 78. The Table \ref{tab:feats} summarizes the main features of the dataset.

\begin{table}[htb!]

\caption{Main features of the dataset. Where [29,28,27], respectively is the number of examples of classes {\itshape Cartão de Crédito e Crédito Rotativo}, {\itshape Registrato} and {\itshape Crédito Imobiliário}.}
\centering\centering\resizebox{1\textwidth}{!}{
\begin{tabular}{|l|c|c|c|}
\hline
\textbf{Column }     & \textbf{Avg. Num. Words }  &  \textbf{Num. Unique }  & \textbf{Top3 Num. Samples } \\ 
\hline
\texttt{Question }   & 12                        & 1855                   & - \\
\hline
\texttt{Category }   & 6                         & 242                    & [29,28,27] \\
\hline
\texttt{Answer }   & 78                        & 1848                   & - \\
\hline
\end{tabular}
}
\label{tab:feats}
\end{table}

\section{Data Augmentation}
\label{sec:data_aug}

Data augmentation (DA) constitutes a classical technique within the realm of machine learning, finding extensive application, particularly in the sub-discipline of computer vision \cite{computer-vision_AUG}. In the domain of computer vision, operations such as rotation, inversion, and discoloration are conventionally applied to images to enhance the inherent diversity of the original data. This methodology proves instrumental in mitigating the challenges associated with limited data availability for model training, facilitating the generation of artificial data to foster improved adaptation and generalization of machine learning models to the targeted problem domain.

The extension of the DA technique to NLP introduces additional intricacies compared to its application in computer vision. The discrete nature of textual data introduces complexities, as alterations to a sentence, such as word substitutions, can profoundly influence sentiment, potentially altering the model's interpretation and subsequent performance. Finally, the work presented in \cite{dataaug} delineates the categorization of DA techniques in NLP into three distinct classes: paraphrasing, noise injection, and sampling.

%DA is a classic technique in computer vision \cite{computer-vision_AUG}, rotating, inverting and discoloring images are some of the several techniques used. In addition to solving the problem of shortage training data, the DA increases the data diversity providing better data generalization and reduces the overfitting issue. DA in NLP has some additional complexities when compared to computer vision. From the discrete nature of text, creating synthetic texts from existing data can be disastrous, the change of a single word in a sentence could change the sentiment of a sentence from negative to positive and negatively affect the model performance. According to \cite{dataaug}, DA methods are divided into three categories: paraphrase, noise, and sampling.

Within this study, DA is harnessed, specifically employing paraphrasing techniques, to augment the textual diversity inherent in the original question-answer pairs sourced from the BACEN FAQ dataset. The transformations applied to the initial dataset involved synthetic replication with alterations to the original texts, ensuring a controlled semantic variance between the synthetic and original texts. This approach aligns with the inherent characteristics of FAQs, where it is posited that queries from distinct users seeking the same answer will exhibit high semantic similarity. This rationale underscores the adoption of the paraphrasing method within the framework of the FAQ dataset.

%In this article, we use made DA by paraphrase: the synthetic copies are slightly modified from the existing data, that is, the semantic difference between the synthetic and the original text is limited. This method was adopted given the nature of the FAQ. We believe that two users looking for the same answer will  ask questions with high semantic similarity.

\subsection{DA Framework on BACEN FAQ}
%\subsection{DA Framework}

Two distinct data augmentation (DA) methodologies were employed in this study, distinguished by their application at the word and sentence levels. The augmentation exclusively targeted the question texts within the training set partition. In the first method, transformations were selectively applied solely to the question-related portions of the dataset. Notably, irrelevant words within a sentence were identified and removed, with a replacement chosen from synonyms. It is noteworthy that altering multiple words within a single sentence may introduce grammatical inaccuracies, compromising the syntactic integrity of the augmented sentence, despite preserving high semantic similarity to the original counterpart. To circumvent the introduction of noise, a constraint was imposed, allowing only one word to be changed per sentence. Consequently, for each original question in the dataset, a corresponding synthetically augmented question at the word level was generated. The second method of DA employed a back-translation process, encompassing translation into English and subsequent retranslation into Portuguese. The T5 model \cite{T5}, configured as per the original specifications in \citeonline{pesoT5}, was utilized for translating questions into English. Subsequently, leveraging the Pegasus model \cite{pegasus}, 10 new texts were generated for each question from the translated corpus. Following this, the newly created set of questions were back-translated to Portuguese. Post-process, redundant sentences were removed, resulting in a dataset expansion of 9.6 times the original volume.

%We have adopted two DA methods, word-level and phrase-level.  The DA was made only in the question texts in the training set split. The main advantage of using word-level DA is simplicity. In our case, we remove the stop-words from a sentence and draw a word to be replaced by a synonym. However, if many words of the same sentence are changed, could make the augmented sentence grammatically incorrect, even if it has high semantic similarity with the original sentence. In order to avoid creating noisy data, we limited the change of one word per sentence, so for each original question in training set split, a synthetic question by word-level was created.

%In sentence-level DA we use the following strategy: we translate the questions into English language with T5 model \cite{T5} and weights from Facebook AI \cite{pesoT5}. From the translated texts, 10 paraphrases were created for each text by using the Pegasus model \cite{pegasus}, then we made {\itshape backtranslated} to the original language. At the end of the process, repeated sentences were eliminated, and we get a 9.6x amount data.

In order to create a three DA objects from training set questions, with the same size as the original questions, we create the embeddings of all DA and measure the similarity by cosine, where we get:

\begin{itemize}
    \item {\bfseries DA$_{\text{SYNONYM}}$}  paraphrase created with word-level, similarity range (0.75, 0.999); 
    \item {\bfseries DA$_{\text{MAX\_SIM}}$} examples chosen with greater similarity of the sentence-level paraphrase with the original question, similarity range between (0.658, 0.996);
   \item {\bfseries DA$_{\text{MIN\_SIM}}$} examples chosen with less similarity of the sentence-level paraphrase with the original question, similarity range between (0.596, 0.965).
\end{itemize} 

\begin{figure}[htb!]
\centering
\includegraphics[width=0.9\linewidth]{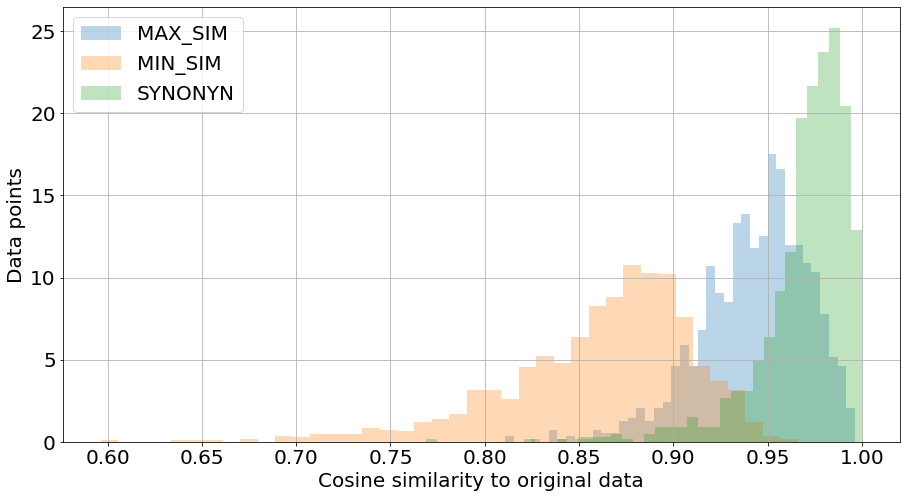} 
\caption{Histogram of DA datasets.}
\label{fig:da-dist}
\end{figure}

The Figure~\ref{fig:da-dist} shows the histogram of cosine similarity of the DA datasets and the Table \ref{tab:aug} shows some samples of DA. %%%% (ACHO QUE poderiamos colocar o score de coseno para cada um dos exemplos na tabela.)

\begin{table}[htb!]
\caption{Samples of DA questions.}
\centering{
\begin{tabular}{|c|c|c|c|c|}
\hline
{\bfseries Question } & {\bfseries DA$_{\text{SYNONYM}}$ } & {\bfseries DA$_{\text{MAX\_SIM}}$ } & {\bfseries DA$_{\text{MIN\_SIM}}$  }\\ 
\hline
% {\itshape A taxa de câmbio é}  & {\itshape A taxa de câmbio}    & {\itshape A taxa de}       & {\itshape A taxa de câmbio}   \\
% {\itshape fixada pelo}         & {\itshape é determinada?}      & {\itshape câmbio é fixa?}  & {\itshape é estabelecida?} \\
% {\itshape Banco Central?}      & {\itshape pelo Banco Central?} &                            & pelo Banco Central? \\
% \midrule
{\itshape Como faço um PIX? }   & {\itshape Como posso}          & {\itshape O que eu faço}   & {\itshape Porque faço} \\
                               & {\itshape fazer um PIX? }       & {\itshape para fazer um PIX? } & {\itshape um PIX? } \\
\hline
{\itshape Como iniciar }        & {\itshape Como começar }        & {\itshape Como começar? }  & {\itshape Como originar }\\
{\itshape a declaração? }       & {\itshape a declaração? }       &                           & {\itshape a declaração? } \\
\hline
{\itshape O que é}        & {\itshape O que é}             & {\itshape Qual a relação} & {\itshape O que é}\\
{\itshape consórcio?}     & {\itshape o Consórcio?}        & {\itshape entre as partes?} & {\itshape coalizão?} \\
\hline
\end{tabular}
}
\label{tab:aug}
\end{table}

\section{Proposed tasks}

The evaluation of the BACEN FAQ dataset will be conducted through a triad of tasks designed to underscore its quality and ascertain the optimal combination of model and data augmentation (DA) techniques for maximizing individual task outcomes. These intrinsic evaluation experiments, centering on traditional Natural Language Processing (NLP) tasks—namely textual classification, semantic search, and FAQ retrieval—are delineated in subsequent sections.

The initial supervised textual classification task involves the examination of a test dataset comprising 400 examples distributed across 241 classes. Subsequently, the second task, semantic search, requires the retrieval of the original question, category, and answer given a synthetically generated question and its original counterpart. The third task, an unsupervised FAQ retrieval scenario, necessitates identifying the correct answer from all possible options when presented with an input question.

\textbf{Models Employed:} In our experimental framework, the Information Retrieval (IR) model encompasses BM25+ \cite{bm25plus}, denoting a modified BM25 model featuring an additional parameter ($\delta$) specifically tailored for scoring long documents. Furthermore, pre-trained models accessible via Hugging Face (HF) are enlisted, including the mBERT multilingual BERT \cite{bert}, BERTIMbau$-$ \cite{bertimbau}, DPR$-$ \cite{dpr} mirroring the size and architecture of BERT, and the BERTaú \cite{bertau} model.

Although the weights for the BERTaú model are currently unavailable on HF, their inclusion in the study is deemed crucial for contextual validation. Trained on chatbot private data focusing on a specific domain, the BERTaú model exhibits enhanced performance within the scope of the original paper \cite{bertau}. However, for the broader scenario delineated in this study, it demonstrates limitations in recognizing numerous words and various types of numbers, resorting to the '[UNK]' (unknown) token when encountering such elements. %%%% Future iterations of the model will involve fresh training incorporating a novel dataset, amalgamating the public dataset BrWaC\footnote{\url{https://huggingface.co/datasets/brwac}} with the anonymized virtual assistant data from a private dataset. This strategic amalgamation is anticipated to endow the model with enhanced generalization capabilities. 
\subsection{Text Classification}
\label{sec:cls}

The textual classification task derived from the BACEN FAQ dataset entailed the assessment of three neural models. Formally, this task involved the evaluation of 400 examples distributed across 241 classes, with the objective of classifying the textual content category based on the corresponding question. Leveraging the data augmentation (DA) training set, comprising 9605 examples, a reduction to 7 classes ensued, each containing fewer than 5 examples per class, with the minimum number of examples for a class set at 6.

The three models developed for this task were mBERT base, the multilingual BERT model \cite{bert}, BERTaú \cite{bertau}, and the DPR model \cite{dpr}. All models shared identical dimensions and architecture and underwent training for 10 epochs. A batch size of 48 samples was employed, and sequences exceeding 32 tokens were subject to truncation. The results, evaluated using the F$_1$ score, are presented in Table~\ref{tab:cls}.

%From 400 examples with 241 classes, we want to classify the category / macro subject text given the question. Due the limited number of training samples (1067 examples) where 136 classes have less than 5 examples per class. We use the DA training set which has 9605 examples. With the DA training set we reduce to 7 classes that have less than 5 examples per class, where the smallest number of examples for a class is 6.

%The experiment had the following configuration: We used three models, the mBERT base the multilingual BERT model \cite{bert}, BERTaú \cite{bertau} and the DPR model \cite{dpr}, all models have the same size and architecture and they were trained by 10 epochs. We used batch size of 48 samples, sequences larger than 32 tokens were truncated. The Table~\ref{tab:cls} displays the results measured using F$_1$ score.

\begin{table}[htb!]
\caption{Classification performance.}
\centering\centering\resizebox{0.6\textwidth}{!}{
\begin{tabular}{|l|c|c|c|}
\hline
{\bfseries Model }              & {\bfseries Question } & {\bfseries Question$_{\text{AUG}}$ }  & {\bfseries Gain \% } \\ 
\hline
\texttt{mBERT}                  & 0.160                  & 0.276                        & 72.5\%  \\
\hline
\texttt{BERTaú}                 & 0.182                  & \textbf{0.314}                                 & 72.5\% \\
\hline
\texttt{BERTimbau}              & 0.161                  & 0.277                                 & 72.0\% \\
\hline
\texttt{DPR$_{\text{QuestionEncoder}}$} & \textbf{0.243} & 0.244                                 & 0.0\%  \\
\hline
\end{tabular}
}
\label{tab:cls}
\end{table}

\begin{figure}[htb!]
\centering
\includegraphics[width=0.8\linewidth]{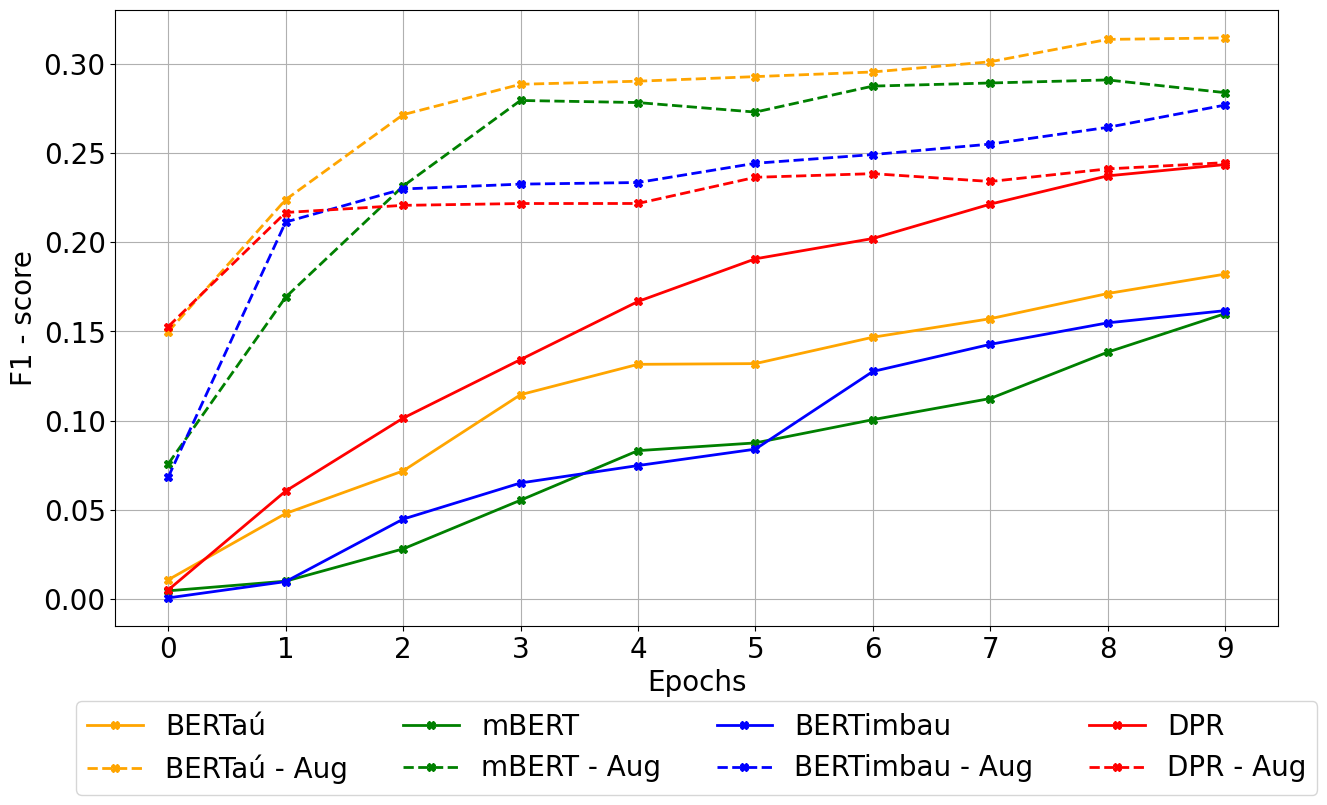} 
\caption{Performance comparison epoch by epoch.}
\label{fig:cls}
\end{figure}

The outcomes presented in Table~\ref{tab:cls} reveal a notable 72\% enhancement in performance attributed to data augmentation (DA) when the BERT model was employed. Conversely, in the case of the DPR model, DA failed to yield any discernible improvement in performance. This observation can be attributed to the intrinsic nature of the DPR training process, where in the optimization of the inner product between pairs of questions and answers emerges as a pivotal factor in achieving optimal performance. For an in-depth understanding of the DPR model \cite{dpr}. A comprehensive, epoch-by-epoch examination is delineated in Figure \ref{fig:cls}.

%For the results shown in Table~\ref{tab:cls}, we observed that the DA contributed positively by 72\% performance when we used BERT model. However, when using the DPR model, we observed that the DA performance did not obtained improvement. We believe that the DPR training process, where the maximization of the internal product of pairs of questions and answers is a key point to the model perform well without DA, for the interested reader in more details of this model, see the original work in \cite{dpr}. A detailed follow-up epoch by epoch is shown in Figure \ref{fig:cls}.

\subsection{Semantic Search}

The semantic search experiment aims to assess the efficacy of data augmentation (DA) datasets in retrieving questions, categories, and answers. To empirically gauge the performance disparity between models with and without DA, we conducted experiments in scenarios both with and without augmentation. The evaluation extends to the quality of category and answer retrieval, with the criteria outlined in \citeonline{portuga-exp} serving as the foundational benchmark.

The test data utilized aligns with that employed in Section~\ref{sec:cls}. The Mean Reciprocal Rank (MRR)@k metric, with k={1.5}, was employed as the evaluation metric given the nature of the ranking task. Leveraging the BERT training weights from the classification task in Section~\ref{sec:cls}, we embedded the texts by aggregating all hidden layers except the first, subsequently conducting cosine similarity assessments between the questions and the target entities. Detailed results are provided in Table \ref{tab:exp2}.

%In this experiment, we are interested in measuring the performance of the DA datasets created to retrieve the question, category and answer. In order to compare the performance of the DA datasets, the performance of the question without DA in the recovery of the category and answer was also investigated, we were inspired to perform this task due to the work of \cite{portuga-exp}. 

%The test dataset is the same used in the \ref{sec:cls} classification task, however we use the Mean Reciprocal Rank (MRR)@k metric with k=\{1,5\} since the problem is treated as a ranking task. We use the BERT's training weights from the classification task \ref{sec:cls}, embedding the texts by adding all the hidden layers, except for the first and performing cosine similarity between the questions and the object to be retrieved: question, category and answer. The results are shown in Table \ref{tab:exp2}.

\begin{table}[htb!]
\centering
\caption{Semantic search performance.}\label{tab:exp2}
\begin{tabular}{|l|c|c|c|c|}
\hline
{\bfseries Model}                     &  {\bfseries MRR@k } & {\bfseries Question } & {\bfseries Category } & {\bfseries Answer }\\
\hline
\texttt{BM25+$_{\text{QUESTION}}$}    & 1                   & -                     & 0.120                 & {\bfseries 0.510} \\
                                      & 5                   & -                     & 0.202                 & {\bfseries 0.602} \\
\hline
\texttt{BM25+$_{\text{SYNONYM}}$}      & 1                  & 0.975                 & 0.102                 & 0.435 \\
                                       & 5                  & 0.985                 & 0.176                 & 0.539 \\
\hline
\texttt{BM25+$_{\text{MAX\_SIM}}$}     & 1                  &  0.917                & 0.080                 & 0.343 \\
                                       & 5                  &  0.940                & 0.141                 & 0.421 \\
\hline
\texttt{BM25+$_{\text{MIN\_SIM}}$}     & 1                  &  0.713                & 0.045                 & 0.250 \\
                                       & 5                  &  0.770                & 0.095                 & 0.312 \\
\hline
\hline
\texttt{BERTaú$_{\text{QUESTION}}$}    & 1                  &  -                    & {\bfseries 0.151}    & {\bfseries 0.482} \\
                                       & 5                  &  -                    & 0.237                 & 0.491 \\
\hline
\texttt{BERTaú$_{\text{SYNONYM}}$}     & 1                  & {\bfseries 0.983}    & 0.123                 & 0.429 \\
                                       & 5                  & {\bfseries 0.994}    & 0.197                 & 0.531 \\
\hline
\texttt{BERTaú$_{\text{MAX\_SIM}}$}    & 1                  &  0.949                & 0.132                 & 0.399 \\
                                       & 5                  &  0.971                & 0.209                 & 0.498 \\
\hline
\texttt{BERTaú$_{\text{MIN\_SIM}}$}    & 1                  &  0.811                & 0.101                 & 0.251 \\
                                       & 5                  &  0.865                & 0.157                 & 0.328 \\
\hline
\hline
\texttt{mBERT$_{\text{QUESTION}}$}     & 1                  &  -                    & 0.147                 & 0.410 \\
                                       & 5                  &  -                    & 0.239                 & 0.501 \\
\hline
\texttt{mBERT$_{\text{SYNONYM}}$}      & 1                  & 0.980                 & 0.122                 & 0.327 \\
                                       & 5                  & 0.983                 & 0.199                 & 0.421 \\
\hline
\texttt{mBERT$_{\text{MAX\_SIM}}$}     & 1                  & 0.927                 & 0.140                 & 0.315 \\
                                       & 5                  & 0.947                 & 0.217                 & 0.398 \\
\hline
\texttt{mBERT$_{\text{MIN\_SIM}}$}     & 1                  & 0.775                 & 0.105                 & 0.252 \\
                                       & 5                  & 0.825                 & 0.175                 & 0.334 \\
\hline
\hline
\texttt{BERTimbau$_{\text{QUESTION}}$} & 1                  &  -                    & 0.140                 & 0.475 \\
                                       & 5                  &  -                    & 0.226                 & 0.583 \\
\hline
\texttt{BERTimbau$_{\text{SYNONYM}}$}  & 1                  &  0.980                & 0.122                 & 0.427 \\
                                       & 5                  &  0.992                & 0.194                 & 0.522 \\
\hline
\texttt{BERTimbau$_{\text{MAX\_SIM}}$} & 1                  &  0.947                & 0.130                 & 0.397 \\
                                       & 5                  &  0.966                & 0.209                 & 0.497 \\
\hline
\texttt{BERTimbau$_{\text{MIN\_SIM}}$} & 1                  &  0.807                & 0.097                 & 0.245 \\
                                       & 5                  &  0.860                & 0.156                 & 0.325 \\
\hline
\end{tabular}
\end{table}

The anticipated superiority of \mbox{BM25+} over BERT in answer retrieval is a logical outcome. Despite utilizing BERT's weights from the classification \mbox{task—optimized} for category \mbox{prediction—BERT} is inherently structured to learn at an effective sentence level, lacking the nuanced multi-word representations crucial for answer retrieval. Consequently, the embeddings of answer representations in BERT may not exhibit semantic salience comparable to those present in BM25. BERT's context-dependent embeddings introduce flexibility for the same word to possess distinct dense representations. For instance, in the sentence: "{\itshape I eat an apple while writing an email on my apple computer}," the cosine similarity between the two occurrences of "{\itshape apple}" is 0.907, deviating from the fixed representation model observed in word2vec \cite{word2vec}.

%where the words maintain a consistent representation (cosine similarity of

%The superior performance of BM25+ compared to BERT in answer retrieval is expected. Even with the BERT's weights from classification task (optimized to predict category), the BERT is designed to learn at effective sentence level, not multi-word representations like those found in answer retrieval. Thus, it is reasonable to assume that the embeddings of the answer representations are not semantically salient as those found in BM25. The BERT's context-dependent embeddings make it more flexible for the same word to have a different dense representation. For example in the sentence: "{\itshape I eat an apple while writing an email on my apple computer.}" the cosine similarity between the two "{\itshape apple}" words is 0.907 and not 1 as in the word2vec \cite{word2vec} model, where the words have a fixed representation.

\subsection{FAQ Retrieval}

The FAQ Retrieval task is inherently unsupervised, lacking labeled data during the training phase. The conventional approach involves assessing the similarity of a query $(q)$ to a set of candidate answers. In our experiment, we construct triplets $(q, a_{+}, a_{-})$, where $(a_{+}, a_{-})$ represent positive and negative answers to the query $(q)$. Optimization is executed through squared L2 distance, serving as a metric for vector similarity. During the training phase, the model endeavors to amplify the vector similarity of the pair $({q, a_{+}})$ while concurrently diminishing the similarity of the negative pair. The model aligns with the principles delineated in the Colbert model \cite{colbert}, and for an in-depth exploration of technical intricacies, we direct the interested reader to the original work.

\subsubsection{Experiment Setup:}
\begin{figure}[htb!]
\centering
\includegraphics[width=0.8\linewidth]{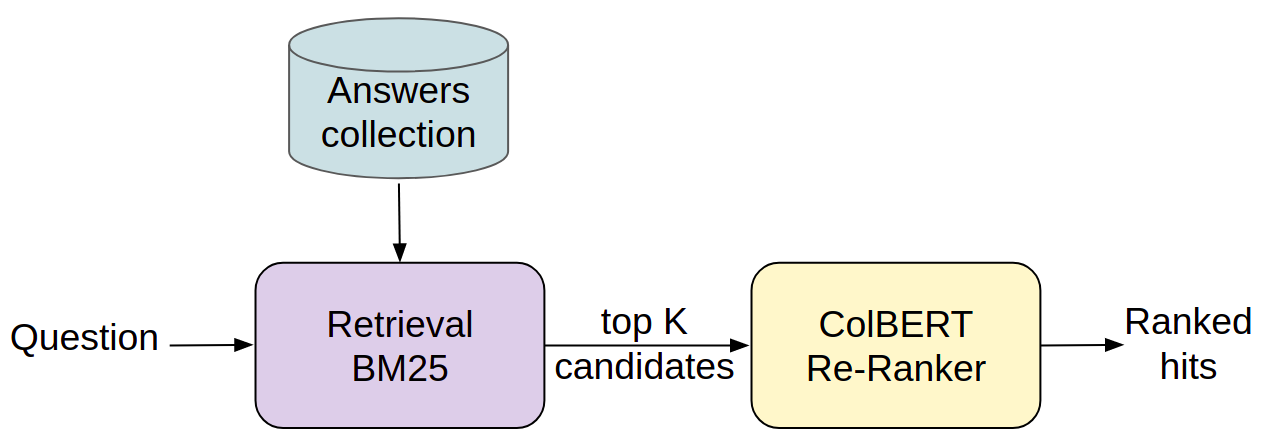} 
\caption{FAQ Retrieval Re-Ranker pipeline.}
\label{fig:reranker}
\end{figure}

We set up the experiment as follows: the train data has 1067 examples, we run one training by 4-epochs for each dataset: no DA, DA$_{\text{SYNONYM}}$, DA$_{\text{MAX\_SIM}}$ and DA$_{\text{MIN\_SIM}}$ and performed the evaluation on the same test dataset as in the previous experiments. Regards the training  strategy: given a question, in the first stage we use the BM25+ to retrieve the top 50 candidates for each question. In a second stage, we use ColBERT framework as a re-ranker that perform attention across the query and the candidate answer and seeks to improve the final results of all first stage candidates. The Figure \ref{fig:reranker} depicts the configuration of the experiment. 

\begin{table}[htb!]
\centering
\caption{FAQ Retrieval performance:  where the numbers in  \text{ColBERT}$_{\text{*}}$ = (1,2,3) are the model weights BERTaú, mBERT e BERTimbau respectively The column Gain was measured with the max score from any ColBERT over the result from BM25. \texttt{Question$_{\text{*}}$=(\texttt{Q, SYN, MAX, MIN})} are the test sets: test, {\bfseries DA$_{\text{SYNONYM}}$}, {\bfseries DA$_{\text{MAX\_SIM}}$} and {\bfseries DA$_{\text{MIN\_SIM}}$} respectively. }\label{tab:exp3}
\begin{tabular}{|l|c|c|c|c|c|c|}
\hline
{\bfseries Model}     & {\bfseries  MRR@k} & {\bfseries BM25+} & {\bfseries ColBERT$_{\text{1}}$}  & {\bfseries ColBERT$_{\text{2}}$} & {\bfseries ColBERT$_{\text{3}}$} & {\bfseries Gain \%}\\
\hline
\texttt{Question$_{\text{Q}}$ }  & 1      & 0.510    & {\bfseries 0.594}    &  0.555  & 0.580   & 16.4\% \\
                                 & 5      & 0.602    & {\bfseries 0.672}    &  0.638  &  0.64   & 11.6\% \\
\hline
\texttt{Question$_{\text{SYN}}$} & 1      & 0.435     & {\bfseries 0.556}   &  0.487  & 0.545   & 27.8\% \\
                                 & 5      & 0.539     & {\bfseries 0.641}   &  0.573  & 0.621   & 18.9\% \\
\hline
\texttt{Question$_{\text{MAX}}$} & 1      & 0.343     & {\bfseries 0.528}   &  0.445  & 0.517   &  {\bfseries 53.9}\% \\
                                 & 5      & 0.421     & {\bfseries 0.608}   &  0.525  & 0.591   &  {\bfseries 44.4}\% \\
\hline
\texttt{Question$_{\text{MIN}}$} & 1      & 0.250     & {\bfseries 0.371}   &  0.347  & 0.357   &  48.4\% \\
                                 & 5      & 0.312     & {\bfseries 0.440}   &  0.418  & 0.429   &  41.0\% \\
\hline
\end{tabular}
\end{table}
We can see in Table \ref{tab:exp3} that re-ranking improves BM25+ performance. Note that as the semantic similarity of the data decreases, i.e,  when the data becomes tricky, the ColBERT gain improves. This result is expected due to the bidirectional architecture of transformers \cite{attention}.

\section{Conflicts of Interest}

Any opinions, findings, and conclusions expressed in this manuscript are those of the authors and do not necessarily reflect the views, official policies nor position of Itaú Unibanco and Central Bank of Brazil.

\section{Conclusion}

During our assessments, data augmentation (DA) yielded improved outcomes in the supervised classification task. However, in the context of the semantic search task, it becomes imperative to scrutinize the linguistic manifestations inherent in DA. The generation of duplicate data with heightened semantic similarity may inadvertently compromise the model's generalization capacity. Notably, we observed that as semantic similarity diminished, the adoption of models featuring contemporary architectures became imperative, surpassing the efficacy of traditional BM25. Nonetheless, a more thorough examination of the outcomes in unsupervised tasks is warranted. Given the delimited domain of financial data, the generation of synthetic data emerges as an increasingly indispensable strategy to meet the escalating demand for Natural Language Processing (NLP) solutions, particularly in scenarios where substantial data volumes remain a prerequisite. 

Our next step is to use a Large Language Model in Portuguese, for example Cabrita \cite{cabrita} to carry out fewshot-learning experiments.

%In our tests, the DA was able to extract better results from the supervised classification task. The semantic search task is necessary to find out which linguistic symptoms the DA carries, creating copies of data with high semantic similarity may imply low power of generalization of the model. We also observed that as the semantic similarity of the data goes down, it was necessary to use models with newer architectures than the BM25. However, further investigation of the results of unsupervised tasks is needed. Given the restricted domain of financial data, the creation of synthetic data is increasingly necessary to attend the growing demand for NLP solutions that still require large amounts of data.
    
%\bibliographystyle{ieeetr}
%\bibliographystyle{IEEEtran}
%\bibliography{bertau}
%\appendix

\bibliographystyle{splncs04}
\bibliography{bertau}

@INPROCEEDINGS{Gori2005, 
author={M. {Gori} and G. {Monfardini} and F. {Scarselli}}, 
booktitle={Proceedings. 2005 IEEE International Joint Conference on Neural Networks, 2005.}, 
title={A new model for learning in graph domains}, 
year={2005}, 
volume={2}, 
number={}, 
pages={729-734 vol. 2}, 
keywords={data structures;graph theory;neural nets;learning (artificial intelligence);graphical data structures;graph neural network;recursive neural networks;learning algorithm;Neural networks;Focusing;Application software;Machine learning;Recurrent neural networks;Encoding;Data structures;Machine learning algorithms;Tree graphs;Software engineering}, 
doi={10.1109/IJCNN.2005.1555942}, 
ISSN={2161-4393}, 
month={July},}

@article{Bengio2009learning,
  title={Learning deep architectures for AI},
  author={Bengio, Yoshua and others},
  journal={Foundations and trends{\textregistered} in Machine Learning},
  volume={2},
  number={1},
  pages={1--127},
  year={2009},
  publisher={Now Publishers, Inc.}
}

@ARTICLE{Hinton2012, 
author={G. {Hinton} and L. {Deng} and D. {Yu} and G. E. {Dahl} and A. {Mohamed} and N. {Jaitly} and A. {Senior} and V. {Vanhoucke} and P. {Nguyen} and T. N. {Sainath} and B. {Kingsbury}}, 
journal={IEEE Signal Processing Magazine}, 
title={Deep Neural Networks for Acoustic Modeling in Speech Recognition: The Shared Views of Four Research Groups}, 
year={2012}, 
volume={29}, 
number={6}, 
pages={82-97}, 
keywords={feedforward neural nets;Gaussian processes;hidden Markov models;speech recognition;deep neural networks;acoustic modeling;speech recognition;hidden Markov models;temporal variability;Gaussian mixture models;feed-forward neural network;posterior probabilities;HMM states;Automatic speech recognition;Speech recognition;Hidden Markov models;Training;Gaussian processes;Acoustics;Neural networks;Data models}, 
doi={10.1109/MSP.2012.2205597}, 
ISSN={1053-5888}, 
month={Nov},}

@inproceedings{Krizhevsky2012imagenet,
  title={Imagenet classification with deep convolutional neural networks},
  author={Krizhevsky, Alex and Sutskever, Ilya and Hinton, Geoffrey E},
  booktitle={Advances in neural information processing systems},
  pages={1097--1105},
  year={2012}
}

@article{Goodfellow2016NIPS2T,
  title={NIPS 2016 Tutorial: Generative Adversarial Networks},
  author={Ian J. Goodfellow},
  journal={CoRR},
  year={2016},
  volume={abs/1701.00160}
}

@book{Blanken2007multimedia,
  title={Multimedia retrieval},
  author={Blanken, Henk M and de Vries, Arjen P and Blok, Henk Ernst and Feng, Ling},
  year={2007},
  publisher={Springer}
}

@inproceedings{Karras2018progressive,
title={Progressive Growing of {GAN}s for Improved Quality, Stability, and Variation},
author={Tero Karras and Timo Aila and Samuli Laine and Jaakko Lehtinen},
booktitle={International Conference on Learning Representations},
year={2018},
url={https://openreview.net/forum?id=Hk99zCeAb},
}

@article{Berthelot2017BEGAN,
  title={BEGAN: Boundary Equilibrium Generative Adversarial Networks},
  author={David Berthelot and Tom Schumm and Luke Metz},
  journal={CoRR},
  year={2017},
  volume={abs/1703.10717}
}

@incollection{Jiajun2016NIPS,
title = {Learning a Probabilistic Latent Space of Object Shapes via 3D Generative-Adversarial Modeling},
author = {Wu, Jiajun and Zhang, Chengkai and Xue, Tianfan and Freeman, Bill and Tenenbaum, Josh},
booktitle = {Advances in Neural Information Processing Systems 29},
pages = {82--90},
year = {2016},
publisher = {Curran Associates, Inc.},
}

@inproceedings{Hamilton2017inductive,
  title={Inductive representation learning on large graphs},
  author={Hamilton, Will and Ying, Zhitao and Leskovec, Jure},
  booktitle={Advances in Neural Information Processing Systems},
  pages={1024--1034},
  year={2017}
}

@article{Kipf2016semi,
  title={Semi-supervised classification with graph convolutional networks},
  author={Kipf, Thomas N and Welling, Max},
  journal={arXiv preprint arXiv:1609.02907},
  year={2016}
} 

@article{Sanchez2018graph,
  title={Graph networks as learnable physics engines for inference and control},
  author={Sanchez-Gonzalez, Alvaro and Heess, Nicolas and Springenberg, Jost Tobias and Merel, Josh and Riedmiller, Martin and Hadsell, Raia and Battaglia, Peter},
  journal={arXiv preprint arXiv:1806.01242},
  year={2018}
}

@inproceedings{Battaglia2016interaction,
  title={Interaction networks for learning about objects, relations and physics},
  author={Battaglia, Peter and Pascanu, Razvan and Lai, Matthew and Rezende, Danilo Jimenez and others},
  booktitle={Advances in neural information processing systems},
  pages={4502--4510},
  year={2016}
}

@inproceedings{Fout2017protein,
  title={Protein interface prediction using graph convolutional networks},
  author={Fout, Alex and Byrd, Jonathon and Shariat, Basir and Ben-Hur, Asa},
  booktitle={Advances in Neural Information Processing Systems},
  pages={6530--6539},
  year={2017}
}

@article{Hamaguchi2017knowledge,
  title={Knowledge transfer for out-of-knowledge-base entities: A graph neural network approach},
  author={Hamaguchi, Takuo and Oiwa, Hidekazu and Shimbo, Masashi and Matsumoto, Yuji},
  journal={arXiv preprint arXiv:1706.05674},
  year={2017}
}

@inproceedings{Khalil2017learning,
  title={Learning combinatorial optimization algorithms over graphs},
  author={Khalil, Elias and Dai, Hanjun and Zhang, Yuyu and Dilkina, Bistra and Song, Le},
  booktitle={Advances in Neural Information Processing Systems},
  pages={6348--6358},
  year={2017}
}

@inproceedings{Bojchevski2018netgan,
  title={NetGAN: Generating Graphs via Random Walks},
  author={Bojchevski, Aleksandar and Shchur, Oleksandr and Z{\"u}gner, Daniel and G{\"u}nnemann, Stephan},
  booktitle={International Conference on Machine Learning},
  pages={609--618},
  year={2018}
}

@article{Zhou2018graph,
  title={Graph Neural Networks: A Review of Methods and Applications},
  author={Zhou, Jie and Cui, Ganqu and Zhang, Zhengyan and Yang, Cheng and Liu, Zhiyuan and Sun, Maosong},
  journal={arXiv preprint arXiv:1812.08434},
  year={2018}
}

@inproceedings{Hinneburg1998,
 author = {Hinneburg, Alexander and Keim, Daniel A.},
 title = {An Efficient Approach to Clustering in Large Multimedia Databases with Noise},
 booktitle = {Proceedings of the Fourth International Conference on Knowledge Discovery and Data Mining},
 series = {KDD'98},
 year = {1998},
 location = {New York, NY},
 pages = {58--65},
 numpages = {8},
 url = {http://dl.acm.org/citation.cfm?id=3000292.3000302},
 acmid = {3000302},
 publisher = {AAAI Press},
 keywords = {clustering algorithms, clustering in multimedia databases, clustering in the presence of noise, clustering of high-dimensional data, density-based clustering},
} 

@article{Zhou2009,
 author = {Zhou, Yang and Cheng, Hong and Yu, Jeffrey Xu},
 title = {Graph Clustering Based on Structural/Attribute Similarities},
 journal = {Proc. VLDB Endow.},
 issue_date = {August 2009},
 volume = {2},
 number = {1},
 month = aug,
 year = {2009},
 issn = {2150-8097},
 pages = {718--729},
 numpages = {12},
 url = {https://doi.org/10.14778/1687627.1687709},
 doi = {10.14778/1687627.1687709},
 acmid = {1687709},
 publisher = {VLDB Endowment},
} 

@article{Scarselli:2009:GNN:1657477.1657482,
 author = {Scarselli, Franco and Gori, Marco and Tsoi, Ah Chung and Hagenbuchner, Markus and Monfardini, Gabriele},
 title = {The Graph Neural Network Model},
 journal = {Trans. Neur. Netw.},
 issue_date = {January 2009},
 volume = {20},
 number = {1},
 month = jan,
 year = {2009},
 issn = {1045-9227},
 pages = {61--80},
 numpages = {20},
 url = {http://dx.doi.org/10.1109/TNN.2008.2005605},
 doi = {10.1109/TNN.2008.2005605},
 acmid = {1657482},
 publisher = {IEEE Press},
 address = {Piscataway, NJ, USA},
 keywords = {Graphical domains, graph neural networks (GNNs), graph processing, graphical domains, recursive neural networks},
} 

@article{DBLP:journals/corr/abs-1805-11724,
  author    = {Michael Kampffmeyer and
               Yinbo Chen and
               Xiaodan Liang and
               Hao Wang and
               Yujia Zhang and
               Eric P. Xing},
  title     = {Rethinking Knowledge Graph Propagation for Zero-Shot Learning},
  journal   = {CoRR},
  volume    = {abs/1805.11724},
  year      = {2018},
  url       = {http://arxiv.org/abs/1805.11724},
  archivePrefix = {arXiv},
  eprint    = {1805.11724},
  timestamp = {Mon, 13 Aug 2018 16:48:00 +0200},
  biburl    = {https://dblp.org/rec/bib/journals/corr/abs-1805-11724},
  bibsource = {dblp computer science bibliography, https://dblp.org}
}

@inproceedings{Zhang:2018:DCC:3178876.3186106,
 author = {Zhang, Yizhou and Xiong, Yun and Kong, Xiangnan and Li, Shanshan and Mi, Jinhong and Zhu, Yangyong},
 title = {Deep Collective Classification in Heterogeneous Information Networks},
 booktitle = {Proceedings of the 2018 World Wide Web Conference},
 series = {WWW '18},
 year = {2018},
 isbn = {978-1-4503-5639-8},
 location = {Lyon, France},
 pages = {399--408},
 numpages = {10},
 doi = {10.1145/3178876.3186106},
 acmid = {3186106},
 publisher = {International World Wide Web Conferences Steering Committee},
} 

@inproceedings{beck-etal-2018-graph,
    title = "Graph-to-Sequence Learning using Gated Graph Neural Networks",
    author = "Beck, Daniel  and
      Haffari, Gholamreza  and
      Cohn, Trevor",
    booktitle = "Proceedings of the 56th Annual Meeting of the Association for Computational Linguistics",
    month = jul,
    year = "2018",
    publisher = "Association for Computational Linguistics",
    pages = "273--283",
}

@inproceedings{Schlichtkrull2018ModelingRD,
  title={Modeling Relational Data with Graph Convolutional Networks},
  author={Michael Sejr Schlichtkrull and Thomas N. Kipf and Peter Bloem and Rianne van den Berg and Ivan Titov and Max Welling},
  booktitle={ESWC},
  year={2018}
}

@inproceedings{ae482107de73461787258f805cf8f4ed,
title = "Spectral networks and locally connected networks on graphs",
author = "Joan Bruna and Wojciech Zaremba and Arthur Szlam and Yann Lecun",
year = "2014",
language = "English (US)",
booktitle = "International Conference on Learning Representations (ICLR2014), CBLS, April 2014",
}

@incollection{NIPS2016_6212,
title = {Diffusion-Convolutional Neural Networks},
author = {Atwood, James and Towsley, Don},
booktitle = {Advances in Neural Information Processing Systems 29},
pages = {1993--2001},
year = {2016},
publisher = {Curran Associates, Inc.},
url = {http://papers.nips.cc/paper/6212-diffusion-convolutional-neural-networks.pdf}
}

@incollection{NIPS2017_6703,
title = {Inductive Representation Learning on Large Graphs},
author = {Hamilton, Will and Ying, Zhitao and Leskovec, Jure},
booktitle = {Advances in Neural Information Processing Systems 30},
pages = {1024--1034},
year = {2017},
publisher = {Curran Associates, Inc.},
url = {http://papers.nips.cc/paper/6703-inductive-representation-learning-on-large-graphs.pdf}
}

@article{Li2016GatedGS,
  title={Gated Graph Sequence Neural Networks},
  author={Yujia Li and Daniel Tarlow and Marc Brockschmidt and Richard S. Zemel},
  journal={CoRR},
  year={2016},
  volume={abs/1511.05493}
}

@inproceedings{tai-etal-2015-improved,
    title = "Improved Semantic Representations From Tree-Structured Long Short-Term Memory Networks",
    author = "Tai, Kai Sheng  and
      Socher, Richard  and
      Manning, Christopher D.",
    booktitle = "Proceedings of the 53rd Annual Meeting of the Association for Computational Linguistics and the 7th International Joint Conference on Natural Language Processing",
    month = jul,
    year = "2015",
    doi = "10.3115/v1/P15-1150",
    pages = "1556--1566",
}

@article{Bahdanau2015NeuralMT,
  title={Neural Machine Translation by Jointly Learning to Align and Translate},
  author={Dzmitry Bahdanau and Kyunghyun Cho and Yoshua Bengio},
  journal={CoRR},
  year={2015},
  volume={abs/1409.0473}
}

@inproceedings{gehring-etal-2017-convolutional,
    title = "A Convolutional Encoder Model for Neural Machine Translation",
    author = "Gehring, Jonas  and
      Auli, Michael  and
      Grangier, David  and
      Dauphin, Yann",
    booktitle = "Proceedings of the 55th Annual Meeting of the Association for Computational Linguistics",
    month = jul,
    year = "2017",
    doi = "10.18653/v1/P17-1012",
    pages = "123--135",
}

@incollection{NIPS2017_7181,
title = {Attention is All you Need},
author = {Vaswani, Ashish and Shazeer, Noam and Parmar, Niki and Uszkoreit, Jakob and Jones, Llion and Gomez, Aidan N and Kaiser, \L ukasz and Polosukhin, Illia},
booktitle = {Advances in Neural Information Processing Systems 30},
pages = {5998--6008},
year = {2017},
publisher = {Curran Associates, Inc.},
url = {http://papers.nips.cc/paper/7181-attention-is-all-you-need.pdf}
}

@inproceedings{cheng-etal-2016-long,
    title = "Long Short-Term Memory-Networks for Machine Reading",
    author = "Cheng, Jianpeng  and
      Dong, Li  and
      Lapata, Mirella",
    booktitle = "Proceedings of the 2016 Conference on Empirical Methods in Natural Language Processing",
    month = nov,
    year = "2016",
    doi = "10.18653/v1/D16-1053",
    pages = "551--561",
}

@article{velickovic2018graph,
  title="{Graph Attention Networks}",
  author={Veli{\v{c}}kovi{\'{c}}, Petar and Cucurull, Guillem and Casanova, Arantxa and Romero, Adriana and Li{\`{o}}, Pietro and Bengio, Yoshua},
  journal={International Conference on Learning Representations},
  year={2018},
  url={https://openreview.net/forum?id=rJXMpikCZ},
}

@INPROCEEDINGS{7780459, 
author={K. {He} and X. {Zhang} and S. {Ren} and J. {Sun}}, 
booktitle={2016 IEEE Conference on Computer Vision and Pattern Recognition (CVPR)}, 
title={Deep Residual Learning for Image Recognition}, 
year={2016}, 
pages={770-778}, 
doi={10.1109/CVPR.2016.90}, 
ISSN={1063-6919}, 
month={June},}

@InProceedings{pmlr-v80-xu18c,
  title = 	 {Representation Learning on Graphs with Jumping Knowledge Networks},
  author = 	 {Xu, Keyulu and Li, Chengtao and Tian, Yonglong and Sonobe, Tomohiro and Kawarabayashi, Ken-ichi and Jegelka, Stefanie},
  booktitle = 	 {Proceedings of the 35th International Conference on Machine Learning},
  pages = 	 {5453--5462},
  year = 	 {2018},
  month = 	 {10--15 Jul},
  publisher = 	 {PMLR},
  url = 	 {http://proceedings.mlr.press/v80/xu18c.html},
}

@inproceedings{chen2018fastgcn,
title={Fast{GCN}: Fast Learning with Graph Convolutional Networks via Importance Sampling},
author={Jie Chen and Tengfei Ma and Cao Xiao},
booktitle={International Conference on Learning Representations},
year={2018},
url={https://openreview.net/forum?id=rytstxWAW},
}

@article{DBLP:journals/corr/abs-1809-05343,
  author    = {Wen{-}bing Huang and
               Tong Zhang and
               Yu Rong and
               Junzhou Huang},
  title     = {Adaptive Sampling Towards Fast Graph Representation Learning},
  journal   = {CoRR},
  volume    = {abs/1809.05343},
  year      = {2018},
  url       = {http://arxiv.org/abs/1809.05343},
  archivePrefix = {arXiv},
  eprint    = {1809.05343},
  timestamp = {Fri, 05 Oct 2018 11:34:52 +0200},
  biburl    = {https://dblp.org/rec/bib/journals/corr/abs-1809-05343},
  bibsource = {dblp computer science bibliography, https://dblp.org}
}

@InProceedings{pmlr-v80-chen18p,
  title = 	 {Stochastic Training of Graph Convolutional Networks with Variance Reduction},
  author = 	 {Chen, Jianfei and Zhu, Jun and Song, Le},
  booktitle = 	 {Proceedings of the 35th International Conference on Machine Learning},
  pages = 	 {942--950},
  year = 	 {2018},
  month = 	 {10--15 Jul},
  publisher = 	 {PMLR},
  url = 	 {http://proceedings.mlr.press/v80/chen18p.html},
  
}

@article{DBLP:journals/corr/GilmerSRVD17,
  author    = {Justin Gilmer and
               Samuel S. Schoenholz and
               Patrick F. Riley and
               Oriol Vinyals and
               George E. Dahl},
  title     = {Neural Message Passing for Quantum Chemistry},
  year      = {2017},
  url       = {http://arxiv.org/abs/1704.01212},
  archivePrefix = {arXiv},
  eprint    = {1704.01212},
}

@article{DBLP:journals/corr/abs-1711-07971,
  author    = {Xiaolong Wang and
               Ross B. Girshick and
               Abhinav Gupta and
               Kaiming He},
  title     = {Non-local Neural Networks},
  journal   = {CoRR},
  volume    = {abs/1711.07971},
  year      = {2017},
  url       = {http://arxiv.org/abs/1711.07971},
  archivePrefix = {arXiv},
  eprint    = {1711.07971},
  timestamp = {Fri, 05 Apr 2019 07:29:46 +0200},
  biburl    = {https://dblp.org/rec/bib/journals/corr/abs-1711-07971},
  bibsource = {dblp computer science bibliography, https://dblp.org}
}

@article{DBLP:journals/corr/abs-1806-01261,
  author    = {Peter W. Battaglia and
               Jessica B. Hamrick and
               Victor Bapst and
               Alvaro Sanchez{-}Gonzalez and
               Vin{\'{\i}}cius Flores Zambaldi and
               Mateusz Malinowski and
               Andrea Tacchetti and
               David Raposo and
               Adam Santoro and
               Ryan Faulkner and
               {\c{C}}aglar G{\"{u}}l{\c{c}}ehre and
               Francis Song and
               Andrew J. Ballard and
               Justin Gilmer and
               George E. Dahl and
               Ashish Vaswani and
               Kelsey Allen and
               Charles Nash and
               Victoria Langston and
               Chris Dyer and
               Nicolas Heess and
               Daan Wierstra and
               Pushmeet Kohli and
               Matthew Botvinick and
               Oriol Vinyals and
               Yujia Li and
               Razvan Pascanu},
  title     = {Relational inductive biases, deep learning, and graph networks},
  journal   = {CoRR},
  year      = {2018},
  url       = {http://arxiv.org/abs/1806.01261},
  archivePrefix = {arXiv},
  eprint    = {1806.01261},
  
}

@article{DBLP:journals/corr/PerozziAS14,
  author    = {Bryan Perozzi and
               Rami Al{-}Rfou and
               Steven Skiena},
  title     = {DeepWalk: Online Learning of Social Representations},
  journal   = {CoRR},
  year      = {2014},
  url       = {http://arxiv.org/abs/1403.6652},
  archivePrefix = {arXiv},
  eprint    = {1403.6652},
  timestamp = {Mon, 13 Aug 2018 16:46:44 +0200},
  biburl    = {https://dblp.org/rec/bib/journals/corr/PerozziAS14},
  bibsource = {dblp computer science bibliography, https://dblp.org}
}

@article{DBLP:journals/corr/GroverL16,
  author    = {Aditya Grover and
               Jure Leskovec},
  title     = {node2vec: Scalable Feature Learning for Networks},
  journal   = {CoRR},
  year      = {2016},
  url       = {http://arxiv.org/abs/1607.00653},
  archivePrefix = {arXiv},
  eprint    = {1607.00653},
  timestamp = {Mon, 13 Aug 2018 16:48:14 +0200},
  biburl    = {https://dblp.org/rec/bib/journals/corr/GroverL16},
  bibsource = {dblp computer science bibliography, https://dblp.org}
}

@article{Kipf2016VariationalGA,
  title={Variational Graph Auto-Encoders},
  author={Thomas N. Kipf and Max Welling},
  journal={CoRR},
  year={2016},
  volume={abs/1611.07308}
}

@article{DBLP:journals/corr/abs-1711-08267,
  author    = {Hongwei Wang and
               Jia Wang and
               Jialin Wang and
               Miao Zhao and
               Weinan Zhang and
               Fuzheng Zhang and
               Xing Xie and
               Minyi Guo},
  title     = {GraphGAN: Graph Representation Learning with Generative Adversarial
               Nets},
  journal   = {CoRR},
  year      = {2017},
  url       = {http://arxiv.org/abs/1711.08267},
  archivePrefix = {arXiv},
  eprint    = {1711.08267},
  timestamp = {Sun, 21 Apr 2019 10:04:41 +0200},
  biburl    = {https://dblp.org/rec/bib/journals/corr/abs-1711-08267},
  bibsource = {dblp computer science bibliography, https://dblp.org}
}

@article{DBLP:journals/corr/LiuCCOYS17,
  author    = {Weiyi Liu and
               Pin{-}Yu Chen and
               Hal Cooper and
               Min Hwan Oh and
               Sailung Yeung and
               Toyotaro Suzumura},
  title     = {Can {GAN} Learn Topological Features of a Graph?},
  journal   = {CoRR},
  year      = {2017},
  url       = {http://arxiv.org/abs/1707.06197},
  archivePrefix = {arXiv},
  eprint    = {1707.06197},
  timestamp = {Mon, 13 Aug 2018 16:47:20 +0200},
  biburl    = {https://dblp.org/rec/bib/journals/corr/LiuCCOYS17},
  bibsource = {dblp computer science bibliography, https://dblp.org}
}

@article{Tavakoli2017,
author = {Tavakoli, Sahar and Hajibagheri, Alireza and Sukthankar, Gita},
year = {2017},
month = {07},
title = {Learning Social Graph Topologies using Generative Adversarial Neural Networks},
doi = {10.13140/RG.2.2.16772.94082}
}

@article{DBLP:journals/corr/YuZWY16,
  author    = {Lantao Yu and
               Weinan Zhang and
               Jun Wang and
               Yong Yu},
  title     = {SeqGAN: Sequence Generative Adversarial Nets with Policy Gradient},
  journal   = {CoRR},
  year      = {2016},
  url       = {http://arxiv.org/abs/1609.05473},
  archivePrefix = {arXiv},
  eprint    = {1609.05473},
  timestamp = {Sun, 21 Apr 2019 10:04:41 +0200},
  biburl    = {https://dblp.org/rec/bib/journals/corr/YuZWY16},
  bibsource = {dblp computer science bibliography, https://dblp.org}
}

@article{articlekusner,
author = {J. Kusner, Matt and Miguel Hernández-Lobato, José},
year = {2016},
month = {11},
pages = {},
archivePrefix = {arXiv},
eprint    = {1611.04051},
title = {GANS for Sequences of Discrete Elements with the Gumbel-softmax Distribution}
}

@article{DBLP:journals/corr/LiMSRJ17,
  author    = {Jiwei Li and
               Will Monroe and
               Tianlin Shi and
               Alan Ritter and
               Dan Jurafsky},
  title     = {Adversarial Learning for Neural Dialogue Generation},
  journal   = {CoRR},
  year      = {2017},
  url       = {http://arxiv.org/abs/1701.06547},
  archivePrefix = {arXiv},
  eprint    = {1701.06547},
  timestamp = {Mon, 13 Aug 2018 16:46:30 +0200},
  biburl    = {https://dblp.org/rec/bib/journals/corr/LiMSRJ17},
  bibsource = {dblp computer science bibliography, https://dblp.org}
}

@article{DBLP:journals/corr/LiangHZGX17,
  author    = {Xiaodan Liang and
               Zhiting Hu and
               Hao Zhang and
               Chuang Gan and
               Eric P. Xing},
  title     = {Recurrent Topic-Transition {GAN} for Visual Paragraph Generation},
  journal   = {CoRR},
  year      = {2017},
  url       = {http://arxiv.org/abs/1703.07022},
  archivePrefix = {arXiv},
  eprint    = {1703.07022},
  timestamp = {Wed, 12 Dec 2018 16:25:50 +0100},
  biburl    = {https://dblp.org/rec/bib/journals/corr/LiangHZGX17},
  bibsource = {dblp computer science bibliography, https://dblp.org}
}

@article{DBLP:journals/corr/CheLZHLSB17,
  author    = {Tong Che and
               Yanran Li and
               Ruixiang Zhang and
               R. Devon Hjelm and
               Wenjie Li and
               Yangqiu Song and
               Yoshua Bengio},
  title     = {Maximum-Likelihood Augmented Discrete Generative Adversarial Networks},
  journal   = {CoRR},
  year      = {2017},
  url       = {http://arxiv.org/abs/1702.07983},
  archivePrefix = {arXiv},
  eprint    = {1702.07983},
  timestamp = {Mon, 13 Aug 2018 16:48:45 +0200},
  biburl    = {https://dblp.org/rec/bib/journals/corr/CheLZHLSB17},
  bibsource = {dblp computer science bibliography, https://dblp.org}
}

@inproceedings{devon2018boundary,
title={Boundary Seeking {GAN}s},
author={R Devon Hjelm and Athul Paul Jacob and Adam Trischler and Gerry Che and Kyunghyun Cho and Yoshua Bengio},
booktitle={International Conference on Learning Representations},
year={2018},
url={https://openreview.net/forum?id=rkTS8lZAb},
}

@article{mccallum2000automating,
  title={Automating the construction of internet portals with machine learning},
  author={McCallum, Andrew Kachites and Nigam, Kamal and Rennie, Jason and Seymore, Kristie},
  journal={Information Retrieval},
  volume={3},
  number={2},
  pages={127--163},
  year={2000},
  publisher={Springer}
}

\begin{thebibliography}{10}
\providecommand{\url}[1]{\texttt{#1}}
\providecommand{\urlprefix}{URL }
\providecommand{\doi}[1]{https://doi.org/#1}

\bibitem{datasets}
Hugging face datasets. \url{https://huggingface.co/docs/datasets/}, accessed: 2021-10-16

\bibitem{bert}
Devlin, J., Chang, M.W., Lee, K., Toutanova, K.: Bert: Pre-training of deep bidirectional transformers for language understanding. Proceedings of the 2019 Conference of the North  (2019). \doi{10.18653/v1/n19-1423}

\bibitem{surveyNLG}
Dong, C., Li, Y., Gong, H., Chen, M., Li, J., Shen, Y., Yang, M.: A survey of natural language generation. ACM Comput. Surv.  \textbf{55}(8) (dec 2022). \doi{10.1145/3554727}, \url{https://doi.org/10.1145/3554727}

\bibitem{pesoT5}
Fan, A., Bhosale, S., Schwenk, H., Ma, Z., El-Kishky, A., Goyal, S., Baines, M., Celebi, O., Wenzek, G., Chaudhary, V., Goyal, N., Birch, T., Liptchinsky, V., Edunov, S., Grave, E., Auli, M., Joulin, A.: Beyond english-centric multilingual machine translation (2020)

\bibitem{bertau}
Finardi, P., Viegas, J.D., Ferreira, G.T., Mansano, A.F., Caridá, V.F.: Berta\'u: Ita\'u bert for digital customer service (2021)

\bibitem{portuga-exp}
Gon{\c{c}}alo~Oliveira, H., Ferreira, J., Santos, J., Fialho, P., Rodrigues, R., Coheur, L., Alves, A.: {AIA}-{BDE}: A corpus of {FAQ}s in {P}ortuguese and their variations. In: Proceedings of the 12th Language Resources and Evaluation Conference. pp. 5442--5449. European Language Resources Association, Marseille, France (May 2020), \url{https://aclanthology.org/2020.lrec-1.669}

\bibitem{dpr}
Karpukhin, V., Oguz, B., Min, S., Lewis, P., Wu, L., Edunov, S., Chen, D., Yih, W.t.: Dense passage retrieval for open-domain question answering. In: Proceedings of the 2020 Conference on Empirical Methods in Natural Language Processing (EMNLP) (Nov 2020)

\bibitem{colbert}
Khattab, O., Zaharia, M.: Colbert: Efficient and effective passage search via contextualized late interaction over bert (2020)

\bibitem{cabrita}
Larcher, C., Piau, M., Finardi, P., Gengo, P., Esposito, P., Caridá, V.: Cabrita: closing the gap for foreign languages  (08 2023)

\bibitem{dataaug}
Li, B., Hou, Y., Che, W.: Data augmentation approaches in natural language processing: A survey (2021)

\bibitem{word2vec}
Mikolov, T., Sutskever, I., Chen, K., Corrado, G.S., Dean, J.: Distributed representations of words and phrases and their compositionality. In: Burges, C.J.C., Bottou, L., Welling, M., Ghahramani, Z., Weinberger, K.Q. (eds.) Advances in Neural Information Processing Systems. vol.~26. Curran Associates, Inc. (2013), \url{https://proceedings.neurips.cc/paper/2013/file/9aa42b31882ec039965f3c4923ce901b-Paper.pdf}

\bibitem{computer-vision_AUG}
Perez, L., Wang, J.: The effectiveness of data augmentation in image classification using deep learning (2017)

\bibitem{T5}
Raffel, C., Shazeer, N., Roberts, A., Lee, K., Narang, S., Matena, M., Zhou, Y., Li, W., Liu, P.J.: Exploring the limits of transfer learning with a unified text-to-text transformer. CoRR  \textbf{abs/1910.10683} (2019)

\bibitem{faq_intro2}
Sakata, W., Shibata, T., Tanaka, R., Kurohashi, S.: Faq retrieval using query-question similarity and bert-based query-answer relevance (2019)

\bibitem{surveyNLP}
Sharifani, K., Amini, M.: Machine learning and deep learning: A review of methods and applications. World Information Technology and Engineering Journal  \textbf{10}(07),  3897--3904 (2023)

\bibitem{bertimbau}
Souza, F., Nogueira, R.F., de~Alencar~Lotufo, R.: Portuguese named entity recognition using {BERT-CRF}. CoRR  \textbf{abs/1909.10649} (2019), \url{http://arxiv.org/abs/1909.10649}

\bibitem{trasfer-learning_1}
Tan, C., Sun, F., Kong, T., Zhang, W., Yang, C., Liu, C.: A survey on deep transfer learning (2018)

\bibitem{bm25plus}
Trotman, A., Puurula, A., Burgess, B.: Improvements to bm25 and language models examined. In: Proceedings of the 2014 Australasian Document Computing Symposium. p. 58–65. ADCS '14, Association for Computing Machinery, New York, NY, USA (2014). \doi{10.1145/2682862.2682863}, \url{https://doi.org/10.1145/2682862.2682863}

\bibitem{attention}
Vaswani, A., Shazeer, N., Parmar, N., Uszkoreit, J., Jones, L., Gomez, A.N., Kaiser, L., Polosukhin, I.: Attention is all you need. CoRR  \textbf{abs/1706.03762} (2017), \url{http://arxiv.org/abs/1706.03762}

\bibitem{pegasus}
Zhang, J., Zhao, Y., Saleh, M., Liu, P.J.: {PEGASUS:} pre-training with extracted gap-sentences for abstractive summarization. CoRR  \textbf{abs/1912.08777} (2019)

\bibitem{faq_intro1}
Zhang, X.F., Sun, H., Yue, X., Lin, S., Sun, H.: Cough: A challenge dataset and models for covid-19 faq retrieval (2021)

\bibitem{trasfer-learning_2}
Zhuang, F., Qi, Z., Duan, K., Xi, D., Zhu, Y., Zhu, H., Xiong, H., He, Q.: A comprehensive survey on transfer learning (2020)

\end{thebibliography}

%\end{thebibliography}
%\bigskip
%\bigskip

\end{document}